\documentclass[final]{cvpr}

\usepackage{times}
\usepackage{epsfig}
\usepackage{graphicx}
\usepackage{amsmath}
\usepackage{amssymb}
\usepackage{kotex}
\usepackage{enumitem}
\setlist[description]{leftmargin=\noindent,labelindent=\noindent}
\usepackage[bottom]{footmisc}

\usepackage[pagebackref=true,breaklinks=true,colorlinks,bookmarks=false]{hyperref}

\newcommand{\floor}[1]{\left\lfloor #1 \right\rfloor}

\begin{document}

\title{Blur More To Deblur Better: Multi-Blur2Deblur For Efficient Video Deblurring}

\author{Dongwon Park$^1$, \quad Dong Un Kang$^1$, \quad Se Young Chun$^*$\\
Department of Electrical Engineering, UNIST, Republic of Korea\\
{\tt\small \{ dong1, qkrtnskfk23, sychun \}@unist.ac.kr}
}

\maketitle

\begin{abstract}
One of the key components for video deblurring is how to exploit neighboring frames. Recent state-of-the-art methods either used aligned adjacent frames to the center frame or propagated the information on past frames to the current frame recurrently. Here we propose multi-blur-to-deblur (MB2D), a novel concept to exploit neighboring frames for efficient video deblurring. Firstly, inspired by unsharp masking, we argue that using more blurred images with long exposures as additional inputs significantly improves performance. Secondly, we propose multi-blurring recurrent neural network (MBRNN) that can synthesize more blurred images from neighboring frames, yielding substantially improved performance with existing video deblurring methods. Lastly, we propose multi-scale deblurring with connecting recurrent feature map from MBRNN (MSDR) to achieve state-of-the-art performance on the popular GoPro and Su datasets in fast and memory efficient ways.
\end{abstract}

\section{Introduction}

\footnotetext[1]{Equal contribution, $^*$Corresponding author}

Video deblurring is a highly \textit{ill-posed} inverse problem that aims to recover the sharp latent image from blurred video frames. The solution for this is getting more and more important due to massive amount of video data from hand-held devices such as smart phones. There are a number of factors to non-uniformly blur videos such as camera shake, object motion, and depth variation that particularly make this inverse problem quite challenging.
Video deblurring is a long-standing computer vision topic. 
Most non-deep learning works investigated how to estimate blur kernels and/or latent frames using neighboring video frames~\cite{chen2008robust,cho2012registration,zhu2012deconvolving,zhang2013multi,zhang2014multi,hee2014gyro}. Among them, multi-image blind deblurring methods have been developed to incorporate observations from multiple blurred images that share the common underlying latent image with different blurs~\cite{chen2008robust,zhu2012deconvolving,zhang2013multi}.

\begin{figure}[!t]
	\centering
	\includegraphics[width=0.9\linewidth]{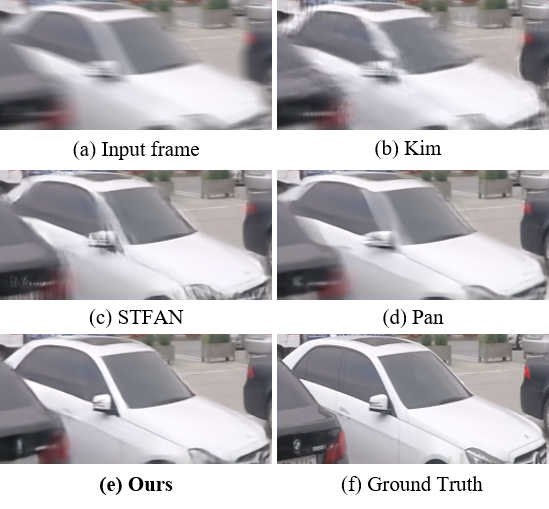}
	\caption{Video deblurring results of 
	Kim~\cite{hyun2017online}, STFAN~\cite{zhou2019spatio}, Pan~\cite{pan2020cascaded}, and our MB2D (Ours) on the GoPro dataset~\cite{nah2017deep}. 
	}
\label{fig:performance}
\vspace{-1em}
\end{figure}

Recent deep learning-based approaches for video deblurring have investigated the ways of utilizing neighboring blurred video frames largely either by using temporally aligned adjacent frames to the reference frame with deep neural networks (DNNs) or by propagating the information about past frames to the reference frame recurrently with recurrent neural network (RNN). 
One group of works exploits adjacent video frames by warping them to the center frame at image and/or feature levels so that sharp pixel information from other frames can be utilized for deblurring~\cite{su2017deep,wang2019edvr,zhou2019spatio,pan2020cascaded}. Due to reduced variations between video frames by alignment, DNNs seem to work more efficiently with aligned frames than with unaltered frames, yielding state-of-the-art (SOTA) video deblurring performance. However, there are a couple of disadvantages for temporal alignment; Alignment in video deblurring itself is also ill-posed and challenging. 
Thus, potential errors in alignment may cause undesirable artifacts in deblurring. Alignment often requires heavy computation cost and large memory for warping operation.
The other group of works utilizes RNNs to sequentially restore the sharp image from video frames using the features from previous steps~\cite{hyun2017online,nah2019recurrent,zhou2019spatio}. Thanks to no alignment operation, these methods have low computation cost, but yielded slightly lower deblurring performance than SOTA methods that were developed around the same time.

Here, we propose multi-blur-to-deblur (MB2D), a novel concept on how to exploit neighboring frames for efficient video deblurring as an alternative to achieve both SOTA performance and fast computation with small memory.

First of all, we propose the fundamental argument for our MB2D: using \emph{more blurred} images with long exposures as additional inputs to video deblurring significantly improves performance. This argument was inspired by classical unsharp masking~\cite{jain1989fundamentals,polesel2000image} that enhances high-frequency components via a more blurred input image. We conjecture that if more blurred images with long exposures for the same reference frame are available, they could encode more information on motions that can potentially improve deblurring performance.
Thus, our proposed MB2D consists of two steps: more-blurring (MB) and then deblurring (D).

\begin{figure}[!t]
	\centering
	\includegraphics[width=0.9\linewidth]{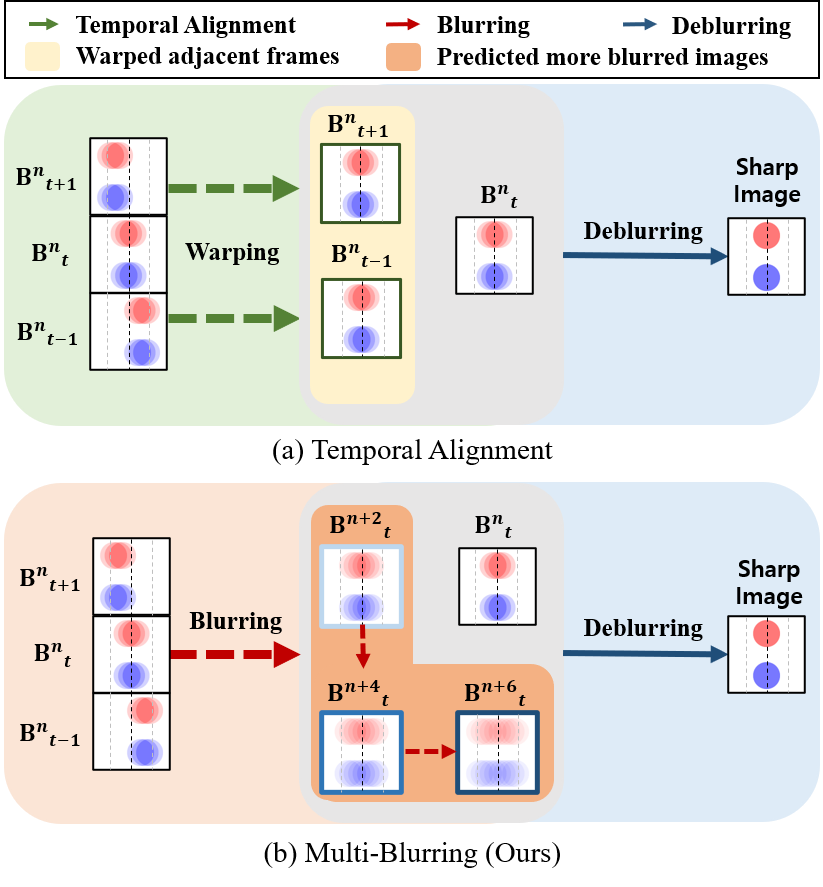}
	\caption{Two pre-processings for video deblurring: (a) Temporal Alignment~\cite{su2017deep,wang2019edvr,zhou2019spatio,pan2020cascaded} that warps neighboring frames to the reference frame ($e.g.$, optical flow) (b) our Multi-Blurring that progressively synthesizes more blurred images with long exposures corresponding to the reference frame using adjacent frames.}
	\label{fig:preprocessing}
	\vspace{-1em}
\end{figure}

Secondly, for the MB step of our MB2D, we propose multi-blurring recurrent neural network (MBRNN) that synthesizes more blurred images (not available during testing) from neighboring blurred frames. Figure~\ref{fig:preprocessing} illustrates (a) video deblurring with alignments~\cite{su2017deep,wang2019edvr,zhou2019spatio,pan2020cascaded} and (b) our MB2D with MBRNN.
While the former warps neighboring frames to the reference frame, the latter (MBRNN) progressively synthesizes \emph{more blurred} images by appending small amounts of motion blurs predicted from neighboring frames.
Our MBRNN is fast and memory-efficient due to no alignment, while yields substantially improved performance with existing video deblurring algorithms.

Lastly, for the D step of our MB2D, we propose multi-scale deblurring with connecting recurrent feature map (CRFM) from MBRNN (MSDR) to achieve SOTA performance for video deblurring on the GoPro and Su datasets in fast and memory efficient ways. Progressively generated recurrent feature maps further improved the performance of video deblurring with the synthesized more blurred images, and thus our proposed MSDR with CRFM performed favorably against existing SOTA methods with relatively small parameters and fast computation as depicted in Figure~\ref{fig:performance}.

The main contributions of this work are summarized as:\\
 \noindent $\bullet$ For the first time, we show that using long exposure images along with the input with regular exposure substantially improves deblurring performance. Then, we propose MB2D, a novel approach with multi-blur and deblur steps.\\
\noindent $\bullet$ We propose MBRNN that progressively synthesizes more blurred images from the input and its neighboring frames. MBRNN improved the performance of existing methods.\\
\noindent $\bullet$ We propose MSDR for video deblurring that exploits CRFM from MBRNN so that the multi-blur and deblur steps are connected at feature levels.\\
\noindent $\bullet$ Our proposed MB2D with MBRNN and MSDR outperformed other SOTA methods on the GoPro~\cite{nah2017deep} and Su~\cite{su2017deep} datasets quickly and parameter-efficiently.

\section{Related Works}

\textbf{Non-DNN multi-images / video deblurring}
Blind deconvolution of motion blur is challenging, 
but multiple blur images often make this inverse problem less \emph{ill-posed}.
Multi-image deblurring has been investigated such as 
multi-channel deconvolution~\cite{zhu2012deconvolving}, 
iterative kernel estimation~\cite{chen2008robust}, multi-channel blind deconvolution with augmented Lagrangian optimization~\cite{sroubek2011robust}, homography estimation for image registration~\cite{cho2012registration}, and multi-image registration~\cite{zhang2014multi}. There have been some works on video deblurring without using DNNs such as patch-based restoration of blurry areas by detecting sharp areas that share the same contents from nearby frames~\cite{cho2012video}, simultaneous estimation of the latent image and optical flow based on locally approximating the pixel-wise varying kernels with bi-directional optical flow~\cite{hyun2015generalized} and local deblurring through weighted Fourier accumulation after warping adjacent frames to the reference by optical flow consistency~\cite{delbracio2015hand}.

\textbf{DNN video deblurring with alignment}
There have been a number of methods for video deblurring to use neighboring frames with temporal alignment. 
Su~\textit{et al.} proposed to align adjacent frames to the center frame, decreasing the spatial variance of video frames~\cite{su2017deep}. EDVR employed deformable CNN to warp the information on adjacent frames to the center frame at feature level~\cite{wang2019edvr}. Zhou~\textit{et al.} proposed STFAN to implicitly estimate the dynamic alignment filters to transform the features in the spatio-temporal filter adaptive network~\cite{zhou2019spatio}. Pan~\textit{et al.} developed the temporal sharpness prior that explores the sharpness pixel and constraints the deep CNN model for generating aligned intermediate latent frames with optical flow~\cite{pan2020cascaded}. 

\textbf{RNN video deblurring}
Kim~\textit{et al.}~\cite{hyun2017online} proposed a method to recurrently estimate the latent image by developing the RNN with the concatenation of multi-frame features. Nah~\textit{et al.}~\cite{nah2019recurrent} developed the RNN to exploit the intra-frame and inter-frame iterations with hidden-state update via reusing RNN cell parameters for video deblurring. Zhong~\textit{et al.}~\cite{zhongefficient} proposed RNN based method to globally aggregate the spatio-temporal correlation from the spatial high-level features of video frames generated by RNN cell. 

\textbf{DNN single image deblurring}
Nah~\textit{et al.}~\cite{nah2017deep} proposed single image deblurring with multi-scale (MS) approach, 
Tao~\textit{et al.}~\cite{tao2018scale} proposed to share models of each stage in the MS approach, and Gao~\textit{et al.}~\cite{gao2019dynamic} developed a partial sharing method considering different level of blurs for each stage in the MS approach. Zhang~\textit{et al.}~\cite{zhang2019deep} and Suin~\textit{et al.}~\cite{suin2020spatially} independently proposed to adjust the receptive field using multi-patch approaches with coarse-to-fine structures. Zhang~\textit{et al.}~\cite{zhang2020deblurring} proposed a data augmentation method by predicting a real blur image using GAN from a single sharp image to reduce the gap between synthetic blur and real blur. Park~\textit{et al.}~\cite{park2019multi} introduced a progressive deblurring method that recurrently estimates the intermediate and final latent images by exploiting time-resolved deblurring data augmentation.

Our MB2D consists of the MB step and the D step. MBRNN utilized RNN, not for deblurring~\cite{hyun2017online,nah2019recurrent}, but for blurring more. 
Unlike~\cite{zhang2020deblurring}, this blurring step is 
for encoding the information of neighboring frames for the same reference. 
MSDR employed the MS approach~\cite{nah2017deep} for deblurring, but with a novel CRFM from MBRNN so that two steps are connected at both image and feature levels.

\section{More Blurred Images with Long Exposures}

This section investigates the fundamental assumption of our proposed MB2D: Using \emph{more blurred} images with long exposures as additional inputs to video deblurring significantly improves performance. More blurred images with long exposures 
are not available during testing, but it is possible to synthesize them during training with video deblur datasets from high-speed cameras~\cite{nah2017deep,su2017deep}. 
While Park~\textit{et al.}~\cite{park2019multi} generated blurred images with short exposures, we propose to synthesize blurred images with long exposures. 

\begin{figure}[!t]
	\centering
	\includegraphics[width=1.0\linewidth]{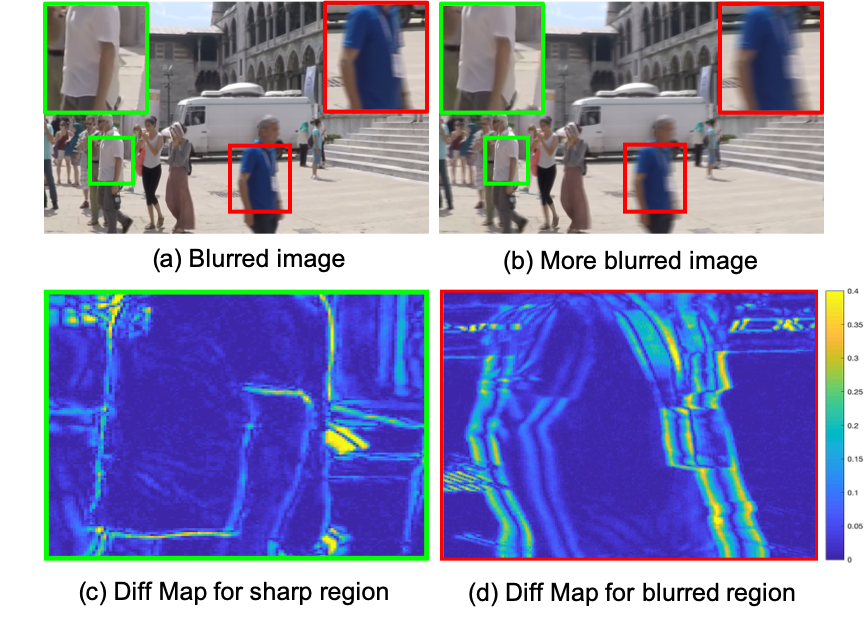}
	\caption{Difference maps between (a) the input blurred image ($B^{11}$) and (b) a more blurred image ($B^{15}$).
   More blurred image extracts high frequency information around edges for sharp regions by small motions (c) and blurred regions by large motions (d).}
	\label{fig:sb_region}
	\vspace{-1em}
\end{figure}

\textbf{Dynamic scenes blur dataset}
Photons are accumulated on a sensor during exposure time, yielding real blurred images. Simulating this, the blurred image $B$  
is generated by the integration of successive sharp images $S$ from high-speed cameras~\cite{nah2017deep,su2017deep} 
as follows: 
\begin{equation}
B^n_t = g\left(\frac{1}{n}\sum_{i=-\floor{n/2}}^{\floor{n/2}}S\left[nt + i\right] \right)
\label{eq:model}
\end{equation}
where $n$ denotes the number of sharp frames (odd, proportional to exposure time for the blurred image $B^n_t$), $S[i]$ denotes the $i$th acquired sharp video frame from a high-speed camera, $g$ is a camera response function and $t$ is an index for the generated blurred image (integer). A typical video deblurring problem is to predict $S[nt]$ or $B^1_t$ (sharp ground truth) from $B^n_t$ and its neighboring blurred frames $B^n_{t-1}$ and $B^n_{t+1}$. Note that as $n$ increases, $B^n_t$ becomes more blurred.

\begin{figure*}[!t]
	\centering
	\includegraphics[width=0.85\linewidth]{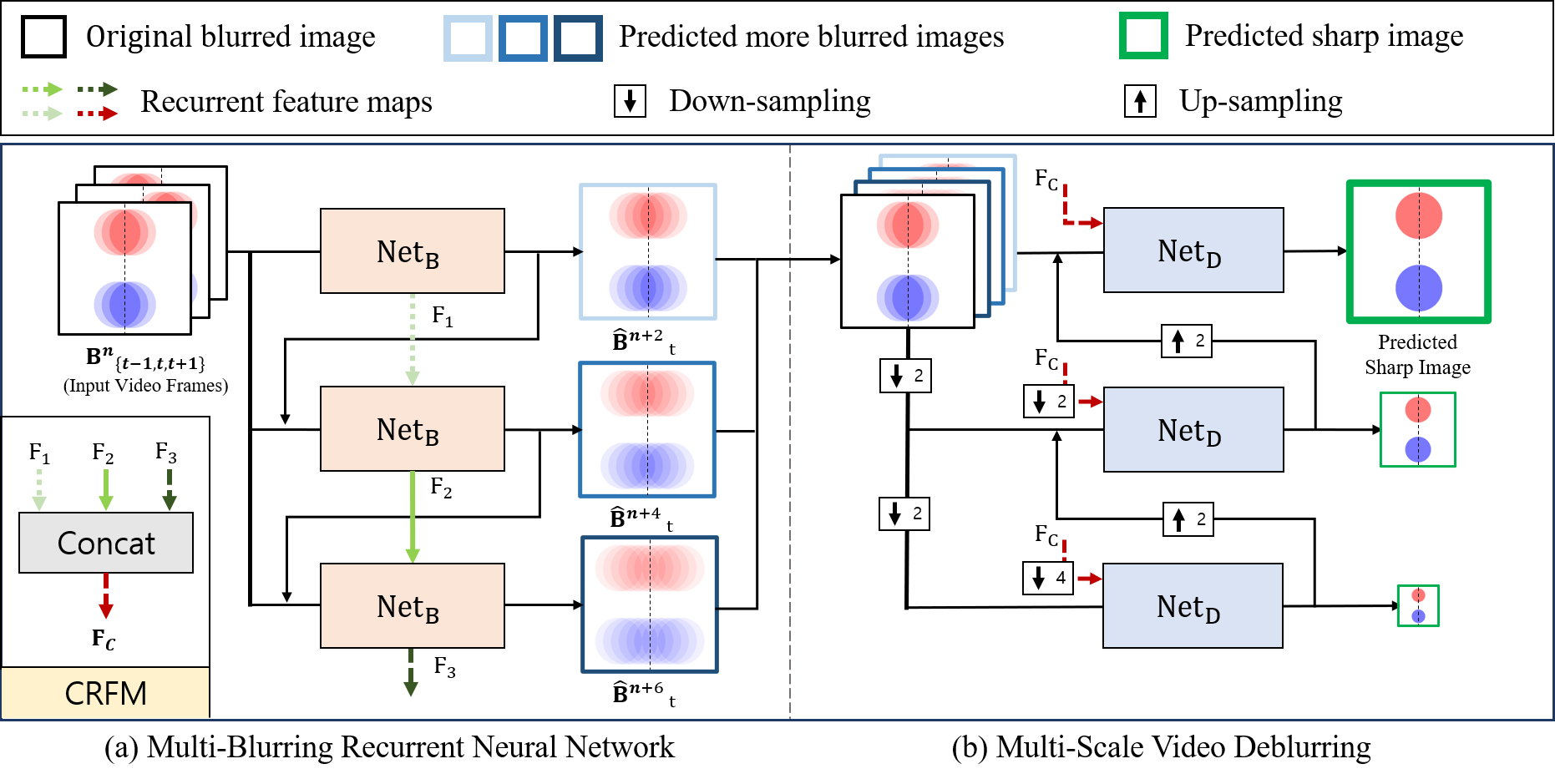}
	\caption{An overview of our MB2D that consists of (a) MB (multi-blurring) and (b) D (deblurring) steps. (a) For given video frames, our MBRNN (denoted by $\mathrm{Net_{B}}$, parameter shared) progressively generates more blurred images by taking previous output and recurrent feature maps. (b) MSDR (denoted by $\mathrm{Net_{D}}$, parameter shared) uses ``coarse-to-fine'' framework with estimated more blurred images from (a) and the input frame. Connecting recurrent feature map from (a) is also used to restore the latent sharp image.}
	\label{fig:process_multi_blurring_approach}
	\vspace{-1em}
\end{figure*}

\subsection{Ideal More Blurred Images}

The unsharp masking~\cite{jain1989fundamentals,polesel2000image} is a classical image sharpening technique to enhance high-frequency components in an image by utilizing the difference between the input image and its blurred (or unsharp) image using Gaussian filtering (we denote this \emph{more blurred image}). This difference emphasizes (or masks) high-frequency components such as edges that were degraded by Gaussian filtering (or more blurring).  
In this work, we extend this idea of unsharp masking to the case of video deblurring.

Instead of Gaussian blurring, we propose to generate more blurred images for video deblurring using sharp images with (\ref{eq:model}). For the blurred image $B^{n}_t$, 
the more blurred video frames for the same reference frame $S[nt]$ will be $B^{n+2}_t$, $B^{n+4}_t$ and $B^{n+6}_t$ that have long exposures to encode more motion information over wide acquisition times as compared to $B^{n}_t$. As illustrated in Figure~\ref{fig:sb_region}, the difference between the input image and a more blurred image encodes the information for sharp region with small motion and blurred region with large motion effectively.

Note that more blurred images can be available during training, but not during testing. Thus, more blurred images are ideal and must be predicted during testing. In addition, for other unsharp masking operations ($e.g.$, scaling), we propose to resort to the power of DNNs (see next Section).

\subsection{More Blurred Images for Deblurring}

We empirically validated our conjecture on more blurred images with deblurring. We performed a single image deblurring with the original input image $B^{11}_{t}$ and/or with ideal more blurred images $B^{13}_{t}$, $B^{15}_{t}$ and so on from the GoPro dataset~\cite{nah2017deep}. We denote the set of input images $B^{11}_{t}, B^{13}_{t}, \ldots, B^{19}_{t}$ by $B^{\{11, 13, \ldots, 19\}}_{t}$. A modified U-Net~\cite{ronneberger2015u} was trained with the input blurred image and/or ideal more blurred images for the same ground truth sharp frame. Single image deblurring results with the ideal more blurred image sets are summarized in Table~\ref{table:effect_multi_images}. Set 1 corresponds to a usual single image deblurring and Sets 2 to 5 correspond to our image deblurring with ideal more blurred images with long exposures.

As compared to the original performance with the input image $B^{11}_{t}$, deblurring with additional more blurred images yielded significantly improved performance. As the more blurred images with longer exposures are used for deblurring, the better performances in PSNR were observed until $n=17$ where dramatic performance boost of 3.33 dB was achieved as compared to the original deblurring with the input with $n=11$. However, after $n=17$, more information from more blurred images did not help to improve deblurring performance. We chose to use 3 more blurred images.

\begin{table}[!b]
\vspace{-1em}
\centering
\caption{Deblurring results with the input and/or ideal more blurred images on the GoPro validation dataset. 
Sets $1, \ldots, 5$ denote $B^{\{11\}}_{t}$, $B^{\{11,13\}}_{t}$, ..., $B^{\{11,13,...,19\}}_{t}$, respectively.}
\begin{tabular}{c|ccccc}
\hline
Set     & 1     & 2     & 3     & \textbf{4}     & 5     \\ \hline 
PSNR(dB) & 29.14 & 31.82 & 32.39 & \textbf{32.47} & 32.39 \\ \hline
\end{tabular}
\label{table:effect_multi_images}
\end{table}

\section{Multi-Blurring To Deblurring (MB2D)}

Figure~\ref{fig:process_multi_blurring_approach} illustrates our proposed MB2D framework that contains the MB step with MBRNN and the D step with MSDR. Our proposed MBRNN in Figure~\ref{fig:process_multi_blurring_approach}(a) aims to progressively predict more blurred images $\hat{B}^{\{n+2,n+4,n+6\}}_{t}$ from the input blurred frame and its neighboring video frames $B^{n}_{\{t-1,t,t+1\}}$ where $B^{n}_{\{t-1,t,t+1\}}$ denotes the set of $B^{n}_{t-1}$, $B^{n}_{t}$, $B^{n}_{t+1}$. Our proposed MSDR in Figure~\ref{fig:process_multi_blurring_approach}(b) uses the input blurred image $B^{n}_{t}$, the predicted more blurred images $\hat{B}^{\{n+2,n+4,n+6\}}_{t}$ as well as the connecting recurrent feature map (CRFM) from MBRNN for multi-scale based video deblurring of estimating the sharp latent image $B^1_t$.

\begin{figure*}[!t]
	\centering
	\includegraphics[width=0.85\linewidth]{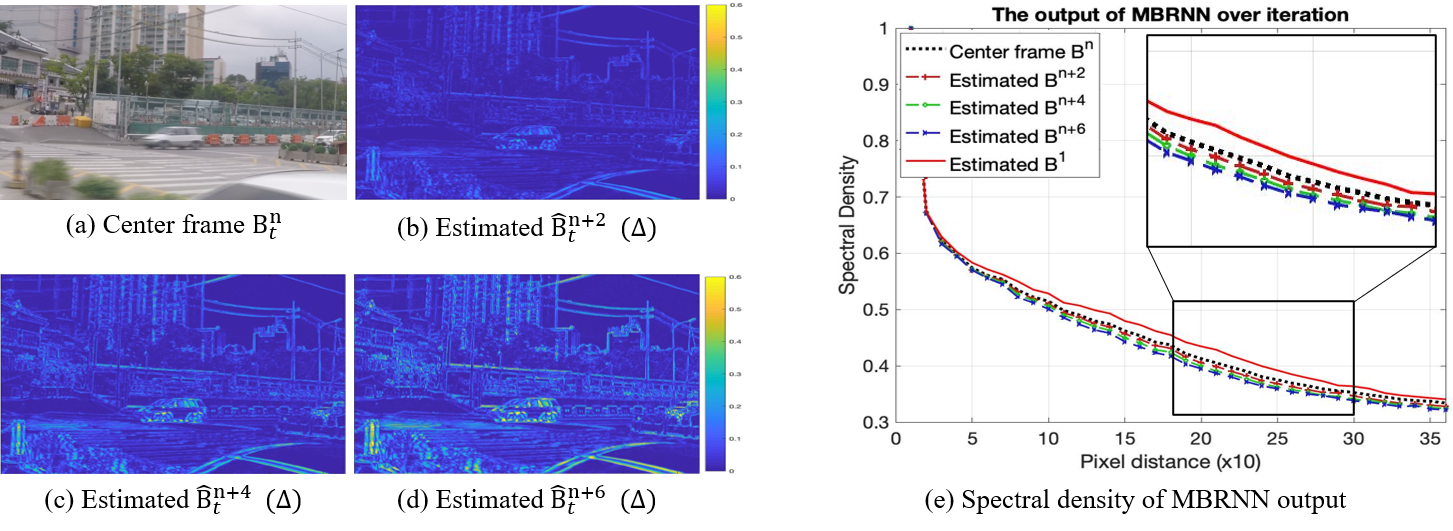}
	\caption{(b-d) Examples of multi-blurrings from MBRNN. $\Delta$ is the absolute difference map between the input center frame (a) and an estimated more blurred image, showing progressively estimated differences by appending small motion blurs to the center frame. (e) Spectral densities of MBRNN outputs, showing progressive degradations of frequency components for more blurred images.}
	\label{fig:blurring_img}
	\vspace{-1em}
\end{figure*}

\subsection{Multi-Blurring Recurrent Neural Network}

Ideal more blurred images significantly improved the performance of deblurring as investigated in Table~\ref{table:effect_multi_images}, but they are available only during training, not during testing.
We conjecture that these more blurred images corresponding to the $n+6$ high-speed video frames can be predicted from the input image $B_t^n$ and its neighboring images $B_{t-1}^n$, $B_{t+1}^n$ that are corresponding to $3n$ high-speed video frames. Thus, the goal of our proposed MBRNN is to progressively predict the multi-blurring images ${B}^{\{n+2,n+4,n+6\}}_{t}$ (available during training) from $B_{\{t-1, t, t+1\}}^n$ (available during training and testing). Our MBRNN is responsible for the MB step of our proposed MB2D.

\textbf{MBRNN model:} A modified U-Net~\cite{ronneberger2015u} was used as the baseline network for our proposed MBRNN for essentially progressive blurring that is similar to the network in~\cite{park2019multi} for progressive deblurring. 
Thus, our proposed MBRNN to yield more blurred images is modeled as:
\begin{equation}
\{ \mathrm{\hat{B}}^{n+2k}_{t}, \mathrm{F}^{k} \} = \mathrm{Net_{B}}(\mathrm{B}^{n}_{\{t-1,t,t+1\}}, \mathrm{\hat{B}}^{n+2k-2}_{t},\mathrm{F}^{k-1})
\end{equation}
where $\mathrm{Net_{B}}$ is the baseline network,
$\mathrm{\hat{B}}^{n+2k}_{t}$ is the estimated more blurred image at the $k$th iteration ($k = 1, 2, 3$), and
$\mathrm{F}^{i}$ is connecting recurrent feature map from the encoder of MBRNN and to the decoder of MBRNN. Figure~\ref{fig:process_multi_blurring_approach}(a) illustrates how the model of MBRNN works to progressively append small amount of motion blurs to the center (or reference) frame by exploiting the estimated more blurred image and intermediate feature map from the previous iteration.
Figure~\ref{fig:blurring_img} shows examples of estimated more blurred images from MBRNN such that 
small motion blurs (b, c, d) are progressively added to the center frame (a), emphasizing fast moving objects. This phenomenon is also observed quantitatively as decreased spectral densities (e) that is the exact opposite to progressive deblurring in~\cite{park2019multi}.

\textbf{MBRNN loss:} 
We trained our proposed MBRNN using a simple L1 function: 
Given $\mathrm{N}$ multi-blurring images $\mathrm{B}^{n+2k}_{t}$ as ground truth that were synthesized from high-speed video frames, the loss function for MBRNN is:
\begin{equation}
\mathcal{L} = \sum_{t=1}^{N}\sum_{k=1}^3 \| \mathrm{\hat{B}}^{n+2k}_{t}  - \mathrm{B}^{n+2k}_{t} \|_{1}
\end{equation}
where $t$ is an index and 
$\mathrm{\hat{B}}^{n+2i}_{t}$ is the output of MBRNN.

\subsection{Multi-Scale Video Deblurring with Connecting Recurrent Feature Map}

Our proposed MB2D consists of the MB step and the D step for more blurring and deblurring, respectively. It is possible to use MBRNN (MB step) and existing video deblurring methods (D step) for improved performance (see next Section for results). While this approach connects the MB and D steps at image level, we propose to connect these two steps at feature levels for further performance boost.

\textbf{MSDR model:} Figure~\ref{fig:process_multi_blurring_approach}(b) illustrates the schematic of the D step using the predicted more blurred images and connecting recurrent feature map from MBRNN. 
Our proposed Multi-Scale video Deblurring network with connecting Recurrent feature map (MSDR) exploits the information from MBRNN, the predicted more blurred images $\hat{B}^{\{n+2, n+4, n+6\}}_{t}$ and the connecting recurrent feature map $F^{(s)}$, in the multi-scale framework for deblurring with down-sampling.
Our proposed MSDR with connecting recurrent feature map is modeled as:
\begin{equation}
\mathrm{\hat{I}}^{(s)} = \mathrm{Net_{D}}({\mathrm{B}_t^{n,(s)}},\mathrm{\hat{B}}^{{\bf m},(s)}_{t},Up(\mathrm{\hat{I}}^{(s+1)}),\mathrm{F}^{(s)})
\end{equation}
where $s$ is a down-scaling index ($s=1,2,3$ for $\times 1, \times 2, \times 4$ downsamplings, respectively),
$\mathrm{\hat{B}}^{{\bf m},(s)}_{t}$ is the set of $\mathrm{\hat{B}}^{n+2,(s)}_{t}$, $\mathrm{\hat{B}}^{n+4,(s)}_{t}$, $\mathrm{\hat{B}}^{n+6,(s)}_{t}$ such that $\mathrm{\hat{B}}^{n+2,(s)}_{t}$ is the downsampled image of $\mathrm{\hat{B}}^{n+2}_{t}$ with the scale index $s$, $\mathrm{\hat{I}}^{(s)}$ is a restored sharp image with the scale index $s$,
$\mathrm{F}^{(s)}$ denotes the downsampled connecting recurrent feature map from MBRNN,
and $\mathrm{Net_{D}}$ is the modified baseline network from $\mathrm{Net_{B}}$ for 
for multi-scale framework (see supplemental for details), and 
$Up$ is a bilinear up-sampling. 
Our MSDR iteratively restores the latent sharp image from down-scales to the original scale.

\textbf{Connecting recurrent feature map:} As illustrated in Figure~\ref{fig:process_multi_blurring_approach}, the recurrent feature maps $\mathrm{F}$ are extracted from the decoders of MBRNN at every iteration and then concatenated before feeding into our MSDR. We call the concatenated feature maps ``connecting recurrent feature map (CRFM)'' that bridges between MBRNN and MSDR at feature levels for the best possible deblurring performance of our proposed MB2D concept.
It seems that CRFM carries more information on blurs around the edges of moving objects than the more blurred images have, leading to the further performance improvement of video deblurring (see 0.64 dB improvement in PSNR via our proposed CRFM as shown in Table~\ref{table:table1}).

\section{Experiments}

\textbf{Datasets}
GoPro dataset~\cite{nah2017deep} consists of 34,874 sharp images (33 videos) with the size of 1280$\times$720 and with 22 videos for training and 11 videos for testing. By the integration of successive sharp images from high-speed cameras, Nah~\cite{nah2017deep} synthesized 2,103 blurred images for training and 1,111 blurred images for testing with sharp ground truth images.
We further generated and added more blurred images to the GoPro dataset (called M-GoPro) for training.

Su dataset~\cite{su2017deep} consists of 6,708 images (71 videos) with the size of 1,920$\times$1080 or 1,280$\times$720 captured by several high-speed camera devices where 61 videos are for training and 10 videos are for testing. 
Following the method of synthesizing motion blurs with interpolated frames using optical flow~\cite{su2017deep}, we further generated and added more blurred images to the Su dataset (called M-Su) for training.

\textbf{Implementation details}
All implementations were done with PyTorch and the Adam optimizer with learning rate $2\times10^{-4}$, $\beta_1=0.9$, $\beta_2=0.999$, $\epsilon=10^{-8}$, and batch size 16 was used as an optimization algorithm for training. Data augmentation techniques for training were used such as random cropping with 256$\times$256 size, random vertical / horizontal flips, and 90$^\circ$ rotation. For Tables~\ref{table:effect_multi_images} to~\ref{table:converting}, the total number of iterations was $10^5$ and for Table~\ref{table:gopro}, it was $10^6$. For Table~\ref{table:su}, the trained network for Table~\ref{table:gopro} with additional fine-tuning of $3\times10^5$ was used.
All experiments were conducted on a NVIDIA Titan V.
Run time was measured with batch size 1 without data-loading time.
MBRNN was trained first and then MSDR was trained with the pre-trained MBRNN.

\begin{table}[!b]
\vspace{-1em}
\centering
\caption{Ablation studies with the number of input frames (NIF) (1 or 3 frames), with pre-processing (Preproc) with optical flow (OF) or our MBRNN (MB), and with our CRFM.}
\label{table:table1}
\begin{tabular}{c c c c l c}
\hline
NIF & Preproc      & CRFM & PSNR   & SSIM & Param(M)    \\ \hline \hline
(a) 1 & -        & -       & 29.29  &0.894     & 2.6    \\
(b) 3  & -        & -       & 29.55  & 0.896    & 2.6    \\
(c) 3  & OF       & -       & 29.63  & 0.901    & 12.0 \\ \hline
(d) 3  & MB & X       & 30.37  & 0.911  & 5.2\\
(e) \textbf{3}  & \textbf{MB} & \textbf{O}       & \textbf{30.94} & \textbf{0.922} & \textbf{5.4} \\ \hline
(f) 3  & - & -       & 30.20 & 0.909 & 5.9 \\
(g) 3  & - & -       & 30.48 & 0.914 & 10.5 \\ \hline
\end{tabular}
\end{table}

\subsection{Ablation Studies for Our MB2D}

We performed the ablation studies to demonstrate the effectiveness of the proposed MB2D approach for several components: the number of input frames (NIF), pre-processing (Pre-proc) such as optical flow (OF) or our MBRNN, and connecting recurrent feature map (CRFM). We chose the modified U-Net as a deblurring network without multi-scale (MS) or multi-temporal (MT) frameworks. The results are summarized in Table~\ref{table:table1}.

Table~\ref{table:table1}(a),(b) correspond to the cases of single image deblurring (with 1 input frame) and of video deblurring (with 3 input frames), respectively, showing the improved performance of (b) over (a) by using more information from neighboring frames.
Table~\ref{table:table1}(b),(c),(d) correspond to different pre-processing methods for video deblurring: Using optical flow for aligned adjacent frames (c) with significantly increased memory requirement (12.0M) improved the deblurring performance by 0.08dB over (b) without pre-processing. However, our proposed MBRNN itself (d) improved the performance by 0.82dB with modestly increased memory requirement (5.2M).
Table~\ref{table:table1}(e) shows additional performance improvement by 0.57dB over (d) when using CRFM from MBRNN for deblurring without increasing network parameters much.

We also performed additional studies on the number of network parameters for (b). As observed in Table~\ref{table:table1}(f),(g), increasing the number of network parameters improved deblurring performance, but our proposed MB2D (especially with CRFM) yielded better performances than the deblurring networks with similar numbers of parameters. Thus, our proposed MB2D yielded improved performance not simply because of increased network size, but because of our proposed approach with more blurred images.

\begin{table}[!b]
\vspace{-1em}
\centering
\caption{Performance comparisons for existing video deblurring approaches (one-stage (OS), multi-temporal (MT), multi-scale (MS)) with/without MBRNN (MB) and CRFM.}
\label{table:ablation}
\begin{tabular}{cccccc}
\hline
    & MB          & CRFM               & PSNR           & SSIM      & Param(M)         \\ \hline \hline
(h) OS          & X         & -                & 29.29          &  0.894    & 2.6          \\
(i) OS          & O         & X                & 30.37          &  0.911    & 5.2          \\
(j) \textbf{OS} & \textbf{O} & \textbf{O} & \textbf{30.94} & \textbf{0.922} & \textbf{5.4} \\ \hline
(k) MT          & X         & -                 & 29.95          &  0.906     & 2.8          \\ 
(l) MT          & O         & X                 & 30.59          &  0.914     & 5.4          \\ 
(m) \textbf{MT} & \textbf{O} & \textbf{O} & \textbf{30.83} & \textbf{0.920} & \textbf{5.6} \\ \hline
(n) MS          & X         & -                 & 29.65          & 0.901         & 2.6          \\ 
(o) MS          & O         & X                 & 30.62          & 0.916         & 5.2          \\ 
(p) \textbf{MS} & \textbf{O} & \textbf{O} & \textbf{31.19} & \textbf{0.926} & \textbf{5.4} \\ \hline
\end{tabular}
\end{table}

Our MB2D concept can be used with various single image / video deblurring approaches such as One-Stage (OS)~\cite{kupyn2019deblurgan,kupyn2018deblurgan,aljadaany2019douglas}, Multi-Temporal (MT)~\cite{park2019multi} and Multi-Scale (MS)~\cite{nah2017deep,tao2018scale,gao2019dynamic} methods where the MS and MT approaches achieve high performance with small number of parameters due to parameter sharing over scales or iterations. We performed another ablation study on them with our proposed MBRNN pre-processing and the results are summarized in Table~\ref{table:ablation}.

Table~\ref{table:ablation}(h),(k),(n) show the results of the baseline methods for OS, MT and MS approaches with the same DNN. In this case, MT approach yielded the best performance among all three approaches by up to 0.66dB. However, with our proposed MBRNN, MS approach yielded the best performance among all approaches by up to 1.08dB as shown in Table~\ref{table:ablation}(i),(l),(o). Further performance improvements were observed with our proposed CRFM from MBRNN and MS approach with them yielded the best performance among all methods by up to 1.54dB. As argued in~\cite{park2019multi}, MS approach seems to destroy high frequency details for good deblurring by using down-sampled images / features, but it seems that our proposed MBRNN and CRFM compensate for the degraded details to yield excellent performance for video deblurring. Thus, we chose MS for the D step of our MB2D.

\begin{table}[!t]
\centering
\caption{Ablation study with different input frames to MBRNN.}
\label{table:analysis}
\begin{tabular}{c|cc|cc}
  & \multicolumn{2}{c|}{1 frame} & \multicolumn{2}{c}{3 frames} \\ \hline 
Output  & PSNR & SSIM & PSNR & SSIM \\ \hline \hline
	$\mathrm{\hat{B}}_{t}^{n+2}$ &    44.73      &     0.994     &     \textbf{46.63}     &     \textbf{0.995}     \\
	$\mathrm{\hat{B}}_{t}^{n+4}$ &     40.89     &      0.987    &     \textbf{43.86}     &     \textbf{0.992}   \\
	$\mathrm{\hat{B}}_{t}^{n+6}$ &     38.54     &      0.980    &     \textbf{42.63}     &     \textbf{0.990} \\\hline
\end{tabular}
\vspace{-1em}
\end{table}

\subsection{Role of Neighboring Frames for MBRNN}

To validate the necessity of adjacent frames to provide motion information for generating more blurred images, we performed an ablation study with 1 frame (without neighboring frames) and 3 frames (with neighboring frames) for MBRNN.
Note that a single frame is generated with $n$ high-speed video frames while 3 frames cover $3n$ high-speed video frames that is usually wider than the more blurred images from $n+2$ to $n+6$ high-speed video frames.
The performance results are summarized in Table~\ref{table:analysis}, showing significant performance differences between the case with a single frame for MBRNN and the case of using neighboring frames. The largest difference was observed for the more blurred image with $n+6$ high-speed video frames and the smallest difference was still significant by 1.9dB. 
Note that MBRNN is computationally efficient (0.02sec) as compared to optical flow (0.076sec)~\cite{sun2018pwc}
that we used for Su~\cite{su2017deep} instead of a MATLAB implementation for it.

\subsection{Multi-Blurring Step for Existing Methods}

\begin{table}[!b]
\vspace{-1em}
\centering
\caption{Performance comparisons of existing methods (Tao~\cite{tao2018scale}, Zhang~\cite{zhang2019deep}, Su~\cite{su2017deep}, Pan\cite{pan2020cascaded}) and their MB2D versions where
$^{*}$ suffix refers to our MB2D version of the existing methods ($e.g.$, Tao$^{*}$ for Tao~\cite{tao2018scale}).
OF is optical flow and MB is our MBRNN.}
\label{table:converting}
\begin{tabular}{cccccc}
\hline
Approach         & Preproc               & PSNR           & SSIM      & Param(M)          \\ \hline \hline
Tao~\cite{tao2018scale}              & -                 &  29.52              &    0.899       & 6.9           \\
\textbf{Tao$^{*}$}     & \textbf{MB} & \textbf{31.12}      & \textbf{0.925} & \textbf{9.7}  \\ \hline
Zhang~\cite{zhang2019deep}          & -                 & 29.91          &  0.905    & 5.4           \\
\textbf{Zhang$^{*}$}   & \textbf{MB} & \textbf{30.85} & \textbf{0.920} & \textbf{8.7}  \\ \hline
Su~\cite{su2017deep}               & OF                & 30.11          & 0.911          & 25.4          \\
\textbf{Su$^{*}$}      & \textbf{MB} & \textbf{30.63} & \textbf{0.915} & \textbf{18.6} \\ \hline
Pan~\cite{pan2020cascaded}          & OF                &30.40                &0.908           &16.19        \\
\textbf{Pan$^{*}$} & \textbf{MB} & \textbf{30.97}      & \textbf{0.923} & \textbf{9.4} \\ \hline
\end{tabular}
\end{table}

Our proposed MB2D is an alternative approach to exploit neighboring frames and its MBRNN can be applied to other existing video deblurring methods with mild modifications such as multi-channel inputs with the predicted more blurred images and with CRFM.
Here, we investigate the feasibility of using our proposed MB2D (especially MBRNN) along with other existing SOTA methods:
Tao~\cite{tao2018scale}, Zhang~\cite{zhang2019deep}, Su~\cite{su2017deep} and Pan~\cite{pan2020cascaded}. 
We implemented our MB2D versions of them, called Tao$^{*}$, Zhang$^{*}$, Su$^{*}$ and Pan$^{*}$, respectively.
Note that for Tao~\cite{tao2018scale}, Zhang~\cite{zhang2019deep}, single image deblurring methods were converted into video deblurring methods with our MB2D by adding MBRNN. For Su~\cite{su2017deep}, Pan~\cite{pan2020cascaded},
we replaced the optical flow network (PWC-Net~\cite{sun2018pwc}) with our MBRNN, resulting in decreased network parameter size by 6.8M. 
Table~\ref{table:converting} shows that our proposed MB2D substantially improved the deblurring performance with existing methods by at least 0.52dB up to 1.6dB for all 4 existing methods.

\begin{table}[!t]
\centering
	\caption{Benchmark qualitative results on the GoPro test dataset~\cite{nah2017deep} for PSNR (dB), SSIM, parameter size (Param in Million) and run time (second).}
	\label{table:gopro}
	\centering
	\begin{tabular}{|l|c|c|c|r|}
		\hline
		Method         & PSNR                  & SSIM                  & Param(M)          &  Time        \\ \hline
		Kim~\cite{hyun2017online}   & 26.82                   & 0.825                & 0.92                       & 0.13      \\
		Su~\cite{su2017deep}   & 27.31                   & 0.826                 & 16.67                      & 2.38      \\
		EDVR~\cite{pan2020cascaded}   & 26.83                   & 0.843                 & -                     & -           \\
        Nah~\cite{nah2019recurrent}  & 29.97					&0.895						 & -                     & -  			\\
		STFAN~\cite{zhou2019spatio}   & 28.59                   & 0.861                 & 5.37                       & 0.15       \\
		Zhong~\cite{zhongefficient}   & 31.07                   & 0.902                 & 1.76                     & 0.54          \\
		Pan~\cite{pan2020cascaded}   & 31.67                   & 0.928                 & 16.19                     & 1.73      \\\hline
		\textbf{Ours}                     & \textbf{32.16}                & \textbf{0.953}              & 5.42                 & 0.27        \\\hline
	\end{tabular}
\end{table}
\begin{table}[!t]
\centering
	\caption{Benchmark qualitative results on the Su test dataset~\cite{su2017deep} for PSNR (dB), SSIM, parameter size (Param in Million) and run time (second).}
	\label{table:su}
	\centering
	\begin{tabular}{|l|c|c|c|r|}
		\hline
		Method         & PSNR                  & SSIM                  & Param(M)          &  Time        \\ \hline
		Kim~\cite{hyun2017online}   & 29.95                   & 0.869                & 0.92                       & 0.13      \\
		Su~\cite{su2017deep}   & 30.01                   & 0.888                 & 16.67                      & 2.38       \\
		EDVR~\cite{pan2020cascaded}   & 28.51                   & 0.864                 & -                     & -           \\
        Nah~\cite{nah2019recurrent}  & 30.80					&0.899						 & -                     & -  			\\
		STFAN~\cite{zhou2019spatio}   & 31.15                   & 0.905                 & 5.37                       & 0.15       \\
		Pan~\cite{pan2020cascaded}   & 32.13                   & 0.927                & 16.19                     & 1.73     \\\hline
		\textbf{Ours}                    & \textbf{32.34}              & \textbf{0.947}              & 5.42                 & 0.27        \\\hline
	\end{tabular}
	\vspace{-1em}
\end{table}

\begin{figure*}[!t]
	\centering
	\includegraphics[width=0.9\linewidth]{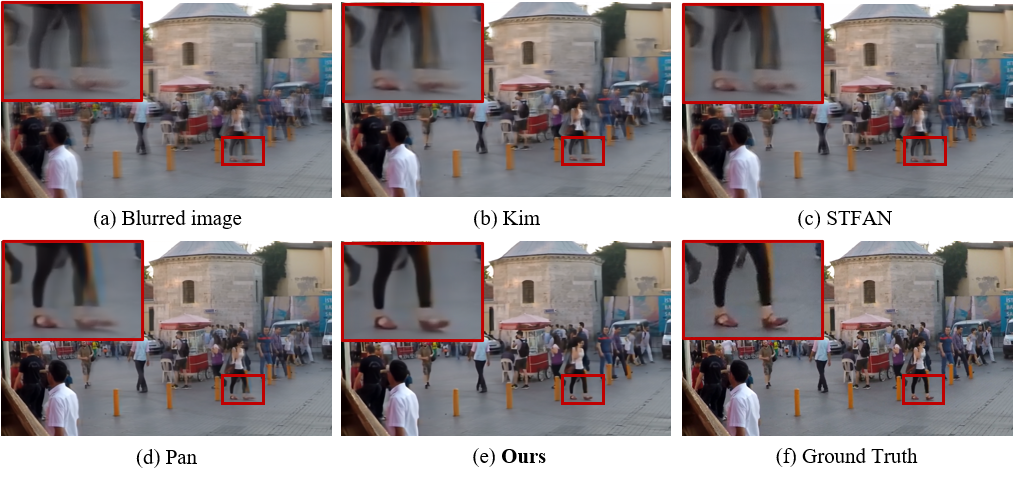}
	\caption{Qualitative results evaluated on the GoPro test dataset~\cite{nah2017deep} for (e) our proposed MB2D method and other SOTA methods: (b) Kim~\cite{hyun2017online}, (c) STFAN~\cite{zhou2019spatio}, and (d) Pan~\cite{pan2020cascaded} for the given input blurred image (a) and its sharp ground truth image (f). Our proposed MB2D yielded visually excellent deblurring results for fast moving persons and objects.}
	\label{fig:benchmark_visual_gopro}
\end{figure*}
\begin{figure*}[!t]
	\centering
	\includegraphics[width=0.9\linewidth]{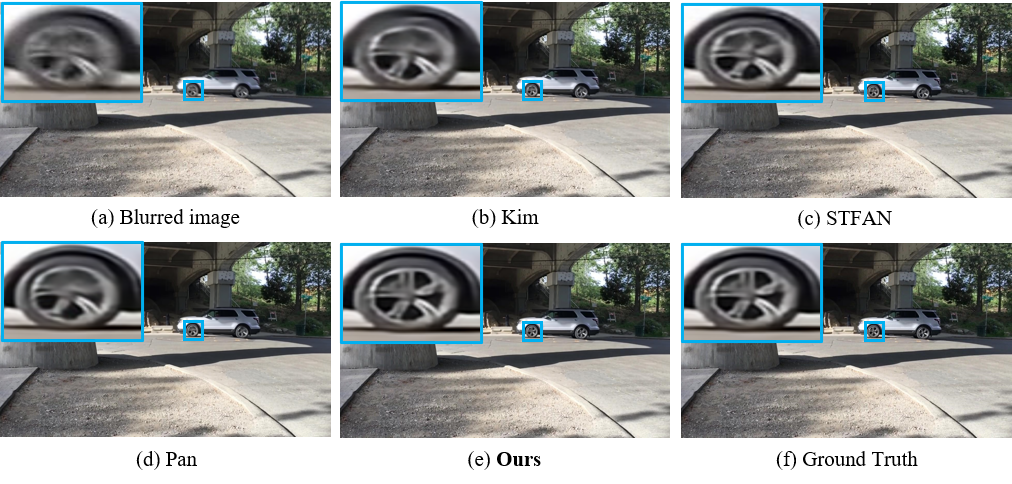}
	\caption{Qualitative results evaluated on the Su test dataset~\cite{su2017deep} for (e) our proposed MB2D method and other SOTA methods: (b) Kim~\cite{hyun2017online}, (c) STFAN~\cite{zhou2019spatio}, and (d) Pan~\cite{pan2020cascaded} for the given input blurred image (a) and its sharp ground truth image (f). Our proposed MB2D yielded visually excellent deblurring results for fast moving persons and objects.}
	\label{fig:benchmark_visual_su}
	\vspace{-1em}
\end{figure*}

\subsection{Benchmark Results}

We performed our proposed MB2D with MBRNN and MSDR on two popular benchmark video deblurring datasets: GoPro test dataset~\cite{nah2017deep} and Su test dataset~\cite{su2017deep}. We use the reported quantitative performances in PSNR (dB) and SSIM of the SOTA methods for Pan~\cite{pan2020cascaded}, Zhong~\cite{zhongefficient} and Nah~\cite{nah2019recurrent}. 
Parameter size and run time results were measured using publicly available codes.
 
Tables~\ref{table:gopro} and~\ref{table:su} show that our proposed MB2D significantly outperformed all existing SOTA methods for video deblurring in terms of PSNR and SSIM on the GoPro and Su test datasets, respectively.
Note that compared to the most recent SOTA method, Pan~\cite{pan2020cascaded}, our proposed MB2D with MBRNN and MSDR yielded substantially outperformed it with more than 3 times less network parameters (5.42M vs. 16.19M) and more than 6 times faster computation time (0.27 second vs. 1.73 second).

Figures~\ref{fig:benchmark_visual_gopro} and ~\ref{fig:benchmark_visual_su} show the examples of video deblurring using a few SOTA methods or using our proposed MB2D on the GoPro and Su test datasets, respectively.
Our proposed MB2D seems to better deblur for fast moving objects or people 
than other existing methods such as Kim~\cite{hyun2017online}, STFAN~\cite{zhou2019spatio}, and Pan~\cite{pan2020cascaded}, 
due to our efficient more blurring based neighboring video frames exploitation.

\section{Conclusion}

We proposed a novel MB2D to exploit neighboring frames for efficient video deblurring. We showed that using more blurred images as additional inputs significantly improves performance, then proposed MBRNN that can synthesize them from neighboring frames, yielding substantially improved performance with existing deblurring methods, and finally proposed MSDR with a novel CRFM to outperform SOTAs in fast and memory-efficient ways.

	\section*{Acknowledgments}
	
	This work was supported partly by 
	Basic Science Research Program through the National Research Foundation of Korea(NRF) 
	funded by the Ministry of Education(NRF-2017R1D1A1B05035810)
	and a grant of the Korea Health Technology R\&D Project 
	through the Korea Health Industry Development Institute (KHIDI), 
	funded by the Ministry of Health \& Welfare, Republic of Korea (grant number: HI18C0316).

{\small
\bibliographystyle{ieee_fullname}
\bibliography{egbib}

\begin{thebibliography}{10}\itemsep=-1pt

\bibitem{aljadaany2019douglas}
Raied Aljadaany, Dipan~K Pal, and Marios Savvides.
\newblock {Douglas-Rachford} networks: Learning both the image prior and data
  fidelity terms for blind image deconvolution.
\newblock In {\em CVPR}, pages 10235--10244, 2019.

\bibitem{chen2008robust}
Jia Chen, Lu Yuan, Chi-Keung Tang, and Long Quan.
\newblock Robust dual motion deblurring.
\newblock In {\em CVPR}, pages 1--8, 2008.

\bibitem{cho2012registration}
Sunghyun Cho, Hojin Cho, Yu-Wing Tai, and Seungyong Lee.
\newblock Registration based non-uniform motion deblurring.
\newblock In {\em CGF}, volume~31, pages 2183--2192, 2012.

\bibitem{cho2012video}
Sunghyun Cho, Jue Wang, and Seungyong Lee.
\newblock Video deblurring for hand-held cameras using patch-based synthesis.
\newblock {\em TOG}, 31(4):1--9, 2012.

\bibitem{delbracio2015hand}
Mauricio Delbracio and Guillermo Sapiro.
\newblock Hand-held video deblurring via efficient fourier aggregation.
\newblock {\em TIP}, 1(4):270--283, 2015.

\bibitem{gao2019dynamic}
Hongyun Gao, Xin Tao, Xiaoyong Shen, and Jiaya Jia.
\newblock Dynamic scene deblurring with parameter selective sharing and nested
  skip connections.
\newblock In {\em CVPR}, pages 3848--3856, 2019.

\bibitem{hee2014gyro}
Sung Hee~Park and Marc Levoy.
\newblock Gyro-based multi-image deconvolution for removing handshake blur.
\newblock In {\em CVPR}, pages 3366--3373, 2014.

\bibitem{hyun2015generalized}
Tae Hyun~Kim and Kyoung Mu~Lee.
\newblock Generalized video deblurring for dynamic scenes.
\newblock In {\em CVPR}, pages 5426--5434, 2015.

\bibitem{hyun2017online}
Tae Hyun~Kim, Kyoung Mu~Lee, Bernhard Scholkopf, and Michael Hirsch.
\newblock Online video deblurring via dynamic temporal blending network.
\newblock In {\em ICCV}, pages 4038--4047, 2017.

\bibitem{jain1989fundamentals}
Anil~K Jain.
\newblock {\em Fundamentals of digital image processing}.
\newblock Prentice-Hall, Inc., 1989.

\bibitem{kupyn2018deblurgan}
Orest Kupyn, Volodymyr Budzan, Mykola Mykhailych, Dmytro Mishkin, and
  Ji{\v{r}}{\'\i} Matas.
\newblock {DeblurGAN:} blind motion deblurring using conditional adversarial
  networks.
\newblock In {\em CVPR}, pages 8183--8192, 2018.

\bibitem{kupyn2019deblurgan}
Orest Kupyn, Tetiana Martyniuk, Junru Wu, and Zhangyang Wang.
\newblock {DeblurGAN-v2:} deblurring (orders-of-magnitude) faster and better.
\newblock In {\em ICCV}, pages 8878--8887, 2019.

\bibitem{nah2017deep}
Seungjun Nah, Tae Hyun~Kim, and Kyoung Mu~Lee.
\newblock Deep multi-scale convolutional neural network for dynamic scene
  deblurring.
\newblock In {\em CVPR}, pages 3883--3891, 2017.

\bibitem{nah2019recurrent}
Seungjun Nah, Sanghyun Son, and Kyoung~Mu Lee.
\newblock Recurrent neural networks with intra-frame iterations for video
  deblurring.
\newblock In {\em CVPR}, pages 8102--8111, 2019.

\bibitem{pan2020cascaded}
Jinshan Pan, Haoran Bai, and Jinhui Tang.
\newblock Cascaded deep video deblurring using temporal sharpness prior.
\newblock In {\em CVPR}, pages 3043--3051, 2020.

\bibitem{park2019multi}
Dongwon Park, Dong~Un Kang, Jisoo Kim, and Se~Young Chun.
\newblock Multi-temporal recurrent neural networks for progressive non-uniform
  single image deblurring with incremental temporal training.
\newblock {\em ECCV}, 2020.

\bibitem{polesel2000image}
Andrea Polesel, Giovanni Ramponi, and V~John Mathews.
\newblock Image enhancement via adaptive unsharp masking.
\newblock {\em TIP}, 9(3):505--510, 2000.

\bibitem{ronneberger2015u}
Olaf Ronneberger, Philipp Fischer, and Thomas Brox.
\newblock U-net: Convolutional networks for biomedical image segmentation.
\newblock In {\em MICCAI}, pages 234--241, 2015.

\bibitem{sroubek2011robust}
Filip Sroubek and Peyman Milanfar.
\newblock Robust multichannel blind deconvolution via fast alternating
  minimization.
\newblock {\em TIP}, 21(4):1687--1700, 2011.

\bibitem{su2017deep}
Shuochen Su, Mauricio Delbracio, Jue Wang, Guillermo Sapiro, Wolfgang Heidrich,
  and Oliver Wang.
\newblock Deep video deblurring for hand-held cameras.
\newblock In {\em CVPR}, pages 1279--1288, 2017.

\bibitem{suin2020spatially}
Maitreya Suin, Kuldeep Purohit, and AN Rajagopalan.
\newblock Spatially-attentive patch-hierarchical network for adaptive motion
  deblurring.
\newblock In {\em CVPR}, pages 3606--3615, 2020.

\bibitem{sun2018pwc}
Deqing Sun, Xiaodong Yang, Ming-Yu Liu, and Jan Kautz.
\newblock {PWC-Net: CNNs} for optical flow using pyramid, warping, and cost
  volume.
\newblock In {\em CVPR}, pages 8934--8943, 2018.

\bibitem{tao2018scale}
Xin Tao, Hongyun Gao, Xiaoyong Shen, Jue Wang, and Jiaya Jia.
\newblock Scale-recurrent network for deep image deblurring.
\newblock In {\em CVPR}, pages 8174--8182, 2018.

\bibitem{wang2019edvr}
Xintao Wang, Kelvin~CK Chan, Ke Yu, Chao Dong, and Chen Change~Loy.
\newblock {EDVR:} video restoration with enhanced deformable convolutional
  networks.
\newblock In {\em CVPRW}, 2019.

\bibitem{zhang2014multi}
Haichao Zhang and Lawrence Carin.
\newblock Multi-shot imaging: joint alignment, deblurring and
  resolution-enhancement.
\newblock In {\em CVPR}, pages 2925--2932, 2014.

\bibitem{zhang2019deep}
Hongguang Zhang, Yuchao Dai, Hongdong Li, and Piotr Koniusz.
\newblock Deep stacked hierarchical multi-patch network for image deblurring.
\newblock In {\em CVPR}, pages 5978--5986, 2019.

\bibitem{zhang2013multi}
Haichao Zhang, David Wipf, and Yanning Zhang.
\newblock Multi-image blind deblurring using a coupled adaptive sparse prior.
\newblock In {\em CVPR}, pages 1051--1058, 2013.

\bibitem{zhang2020deblurring}
Kaihao Zhang, Wenhan Luo, Yiran Zhong, Lin Ma, Bjorn Stenger, Wei Liu, and
  Hongdong Li.
\newblock Deblurring by realistic blurring.
\newblock In {\em CVPR}, pages 2737--2746, 2020.

\bibitem{zhongefficient}
Zhihang Zhong, Ye Gao, Yinqiang Zheng, and Bo Zheng.
\newblock Efficient spatio-temporal recurrent neural network for video
  deblurring.
\newblock {\em ECCV}, 2020.

\bibitem{zhou2019spatio}
Shangchen Zhou, Jiawei Zhang, Jinshan Pan, Haozhe Xie, Wangmeng Zuo, and Jimmy
  Ren.
\newblock Spatio-temporal filter adaptive network for video deblurring.
\newblock In {\em CVPR}, pages 2482--2491, 2019.

\bibitem{zhu2012deconvolving}
Xiang Zhu, Filip {\v{S}}roubek, and Peyman Milanfar.
\newblock Deconvolving {PSF}s for a better motion deblurring using multiple
  images.
\newblock In {\em ECCV}, pages 636--647, 2012.

\end{thebibliography}
}

\end{document}